\documentclass[10pt]{cai}

\begin{document}
\begin{center}

\title{Residual Reinforcement Learning for Waste-Container Lifting Using Large-Scale Cranes with Underactuated Tools}
\maketitle

\thispagestyle{empty}

\begin{tabular}{cc}
Qi Li\upstairs{\affilone,*}, Karsten Berns\upstairs{\affilone}
\\[0.25ex]
{\small \upstairs{\affilone} Robotics Research Lab, RPTU University of Kaiserslautern-Landau, Kaiserslautern, Germany} \\

\end{tabular}

\emails{
\upstairs{*} qili@rptu.de
}
\vspace*{0.2in}
\end{center}

\begin{abstract}
This paper studies the container lifting phase of a waste-container recycling task in urban environments, performed by a hydraulic loader crane equipped with an underactuated discharge unit, and proposes a residual reinforcement learning (RRL) approach that combines a nominal Cartesian controller with a learned residual policy. All experiments are conducted in simulation, where the task is characterized by tight geometric tolerances between the discharge-unit hooks and the container rings relative to the overall crane scale, making precise trajectory tracking and swing suppression essential. The nominal controller uses admittance control for trajectory tracking and pendulum-aware swing damping, followed by damped least-squares inverse kinematics with a nullspace posture term to generate joint velocity commands. A PPO-trained residual policy in Isaac Lab compensates for unmodeled dynamics and parameter variations, improving precision and robustness without requiring end-to-end learning from scratch. We further employ randomized episode initialization and domain randomization over payload properties, actuator gains, and passive joint parameters to enhance generalization. Simulation results demonstrate improved tracking accuracy, reduced oscillations, and higher lifting success rates compared to the nominal controller alone.
\end{abstract}


\section{Introduction}
\begin{figure}[t]
	\centering
	\includegraphics[width=0.8\textwidth]{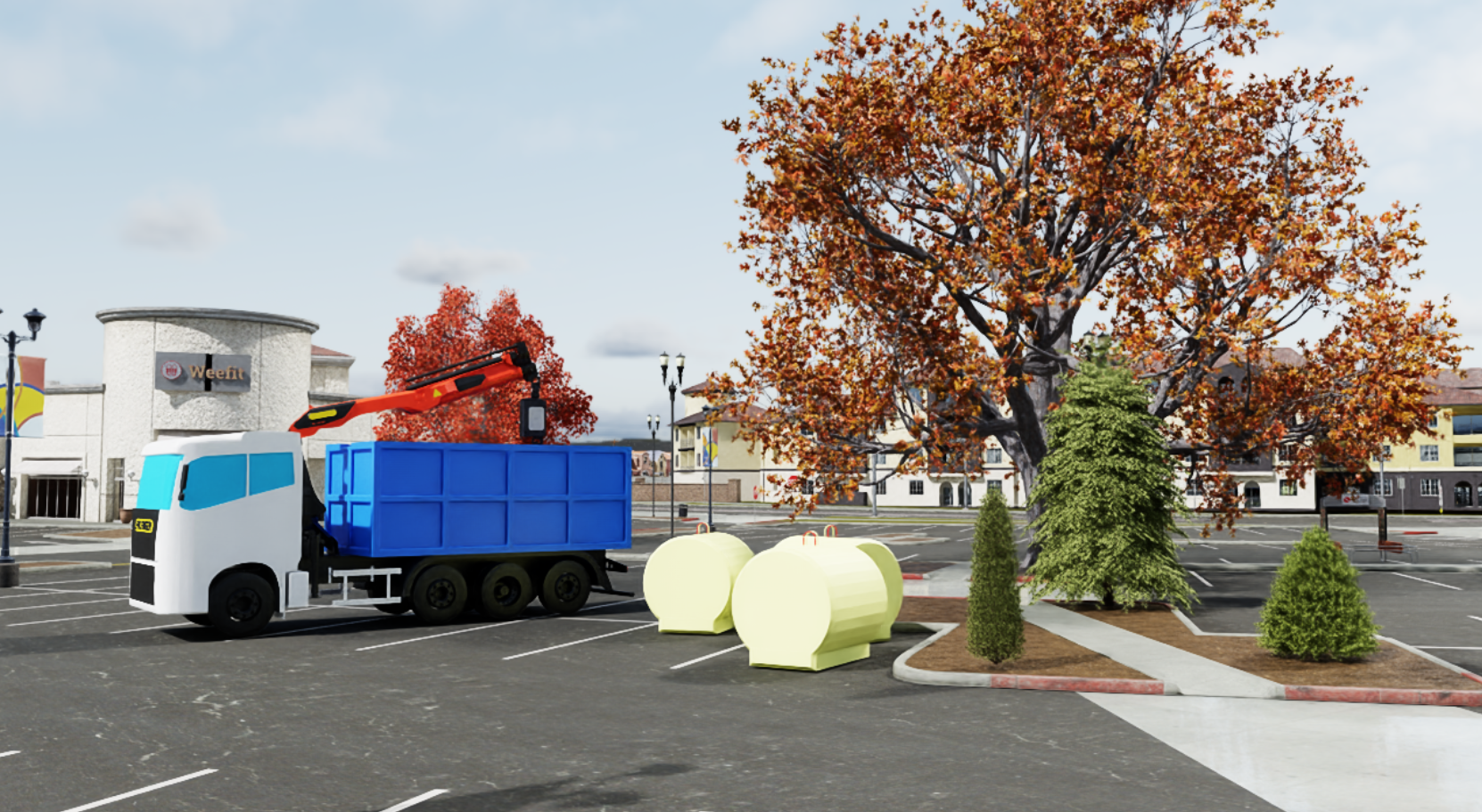}
	\caption{Simulated waste-container handling scenario.}
	\label{fig:scenario}
\end{figure}

Container recycling, such as waste-glass and garbage collection, is an essential component of modern urban infrastructure. As illustrated in Fig.~\ref{fig:scenario}, this task is commonly performed on city streets using a truck-mounted hydraulic loader crane equipped with an underactuated discharge unit. Containers are typically located above or below ground level and are fitted with small hooking rings that must be accurately engaged to lift and empty the container into the truck.
This operation is difficult because (i) cranes are typically commanded in joint space,
(ii) successful hooking requires accurate TCP positioning under tight tolerances, and
(iii) the underactuated discharge unit exhibits oscillations induced by crane motion.
The combination of large-scale crane dynamics, underactuated tools, and small tolerance requirements makes the task physically and cognitively demanding for human operators.
This difficulty is further exacerbated by the shortage of experienced operators and the high cost and time required for training, motivating the need for automation.

Automation of hydraulic machinery has received increasing attention in robotics research, with applications in construction, forestry, recycling, warehousing, and port operations~\cite{Ryan23, andersson2021, Sun2017OffshoreCrane}. While prior work has addressed material handling and lifting with hydraulic systems~\cite{Quang02}, most studies do not focus on high-precision manipulation under tight tolerances. Although accurate manipulation has been extensively studied for industrial robotic arms~\cite{han2023systematic}, hydraulic cranes differ significantly in scale, compliance, actuation, and dynamics, limiting the direct transfer of existing methods.

In this work, we address the problem of accurate container lifting in urban recycling scenarios. We propose a residual reinforcement learning (RRL) framework that combines a reliable nominal Cartesian controller—responsible for trajectory tracking and swing suppression—with a learned residual policy that compensates for unmodeled dynamics and improves precision. This hybrid approach enables accurate and robust control without relying on end-to-end learning from scratch.

This paper is organized as follows. Section~2 reviews related work.
Section~3 presents the system, simulation, and the proposed RRL architecture.
Section~4 reports experiments, and Section~5 provides ablations.
Section~6 concludes and outlines future work.

\section{Related Work}
\subsection{Hydraulic Machinery and Underactuated Loads}
Model-based control of hydraulic manipulators has a long history, addressing nonlinear
actuation and load-dependent dynamics~\cite{Quang02,CHANG2002119,Mattila17}. Integrated
autonomous systems such as HEAP demonstrate full-stack autonomy using classical
controllers~\cite{Jud_2021}, but typically require detailed models and extensive tuning.

Reinforcement learning (RL) has recently been applied to hydraulic equipment, including
excavation subtasks such as bucket filling and soil-adaptive excavation~\cite{Egli24,Egli22},
and scalable training paradigms for excavation~\cite{zhai2025,gruetter2025,Egli20}.
For underactuated crane-like systems, learning-based control has been studied for
multi-pendulum crane dynamics~\cite{Wu24} and highly dynamic motions such as throwing
and material handling~\cite{werner2024,spinelli2025}. Forestry crane RL has also been
reported~\cite{andersson2021,wallin2024}. These efforts highlight strong adaptability,
but they typically do not focus on precise hooking under tight tolerances with large-scale
structural compliance.

\subsection{Residual Reinforcement Learning}
Residual reinforcement learning (RRL) augments a nominal controller with a learned
residual policy to compensate for modeling errors and disturbances. Prior work shows
rapid learning when residual policies refine stable impedance or demonstration-based
controllers~\cite{Kulkarni22,alakuijala2021,ankile2025}. RRL is particularly appealing
for large-scale systems where end-to-end RL can be sample-inefficient. Here, we apply
RRL to combine trajectory tracking and anti-sway control with a residual policy that
improves robustness and precision.

\section{Methodology}

\subsection{System Description and Simulation Setup}
\begin{figure}[t]
	\centering
	\includegraphics[width=1.0\textwidth]{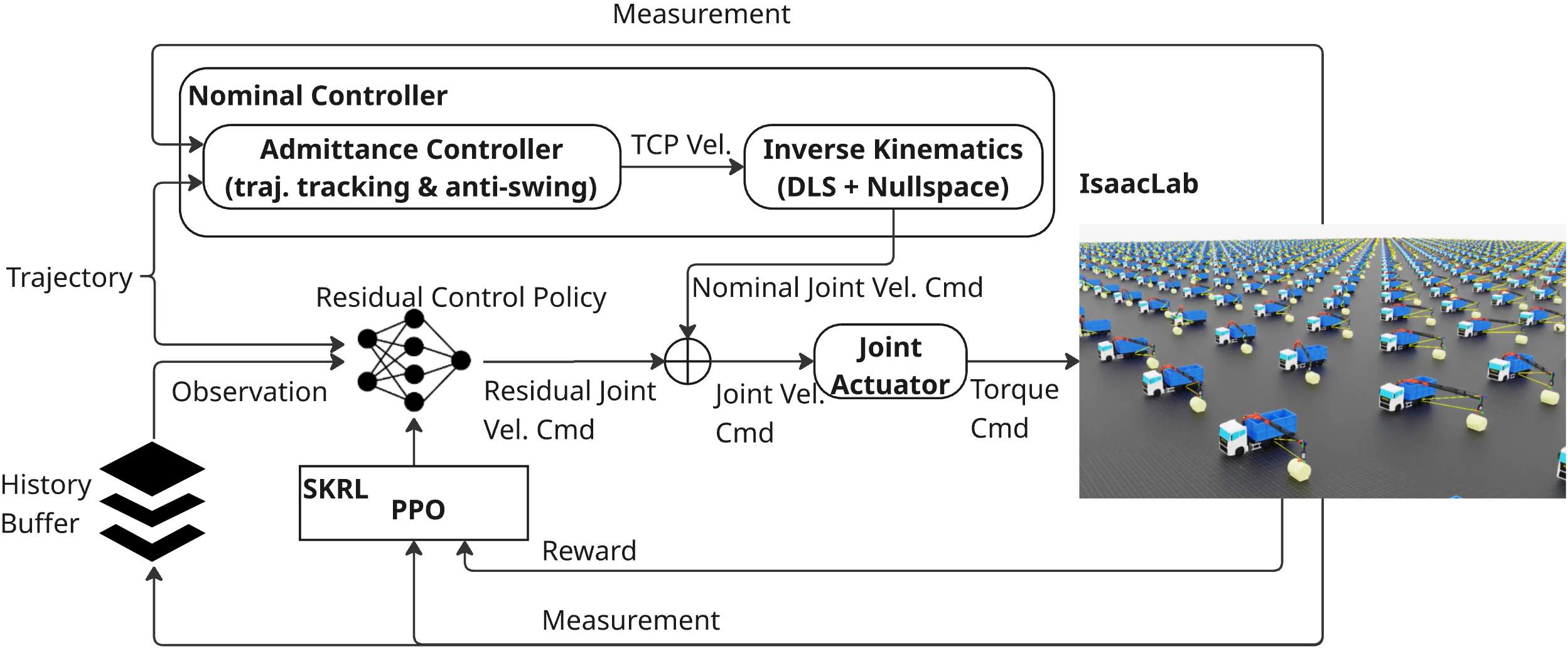}
	\caption{Overall control architecture. The nominal controller outputs \({}^{\mathrm{nor}}u\) (TCP admittance + anti-swing + IK). A PPO-trained residual policy outputs \({}^{\mathrm{res}}u\). The final joint-velocity command is \(u=(1-\lambda){}^{\mathrm{nor}}u+\lambda {}^{\mathrm{res}}u\) with error-dependent blending \(\lambda\), and is applied to joint-level actuators.}
	\label{fig:architecture}
\end{figure}

We consider only the \emph{container lifting} phase of the recycling workflow; transport,
unloading, and release are omitted. All methods are developed and evaluated in Isaac
Lab~\cite{nvidia2025}. The full system is summarized in Fig.~\ref{fig:architecture}. 
The simulated system is based on a modified loader crane mounted on a truck. The truck base is treated as rigid and fixed (no vehicle dynamics). The crane is a 7-DoF serial chain (3 revolute, 4 prismatic), with approximately 13~m outreach and a maximum lifting capacity of 7.1~t.

Hydraulic actuation and linkages are not explicitly modeled. Each joint is driven by a low-level PID controller whose gains are tuned and randomized to reproduce effects similar to structural compliance and bending that occur in large cranes. 
A discharge unit is attached at the crane tool center point (TCP) with two actuated joints (container rotation and hook open/close).
Two \emph{unactuated}
revolute joints between the TCP and the discharge unit model underactuation; their motion
is induced by gravity and crane accelerations, with small friction for damping (similar
in spirit to pendulum modeling but captured directly by rigid-body simulation)~\cite{Spinelli_2024}. The container is equipped with two hooking rings and has dimensions of roughly $1.6 \times 1.6 \times 1.8$~m, with a mass
ranging from 100~kg to 700~kg.

\begin{figure}[t]
	\centering
	\begin{minipage}[t]{0.69\textwidth}
		\vspace{0pt}
		\centering
		\includegraphics[width=\linewidth]{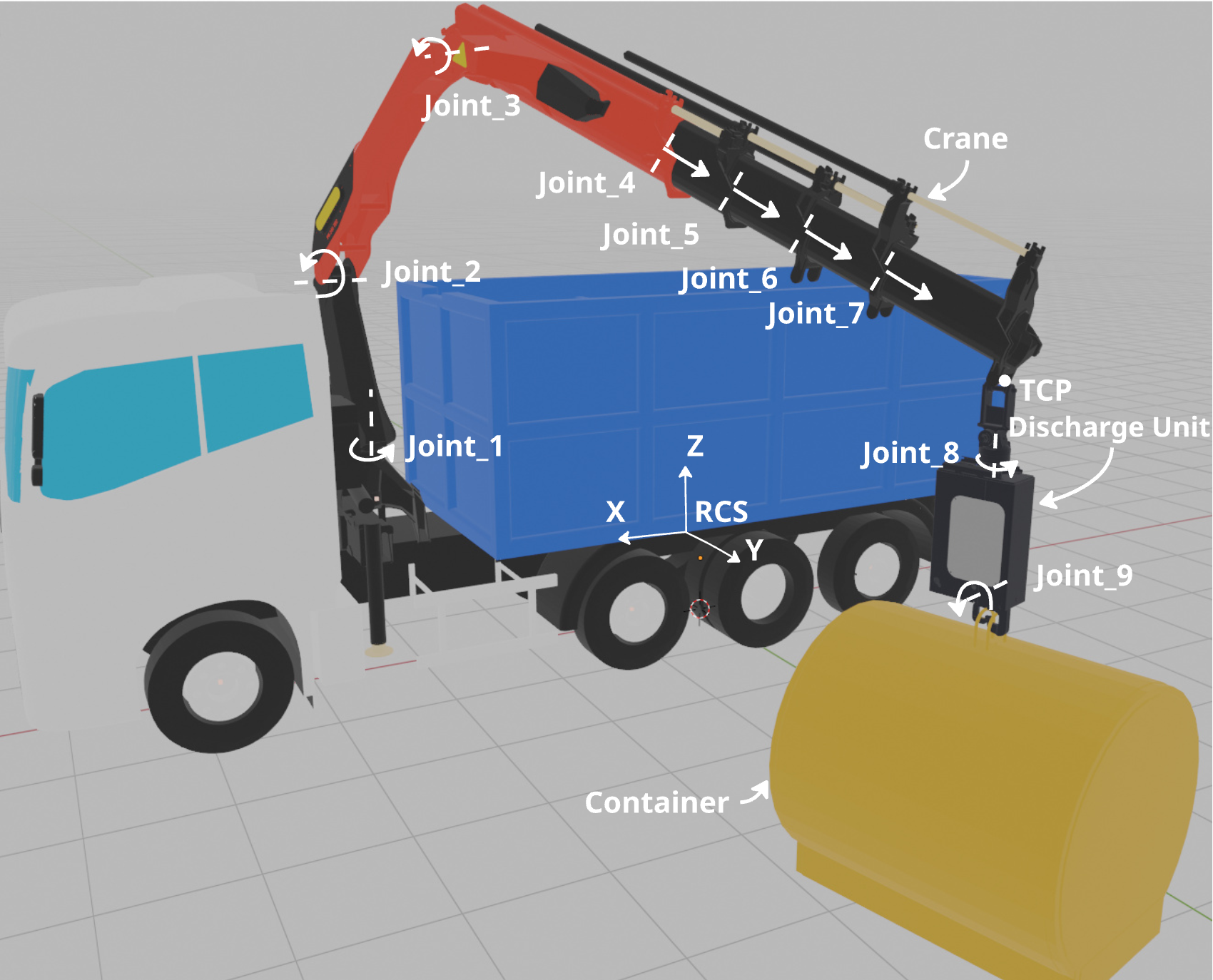}
	\end{minipage}\hfill
	\begin{minipage}[t]{0.296\textwidth}
		\vspace{0pt}
		\centering
		\includegraphics[width=\linewidth]{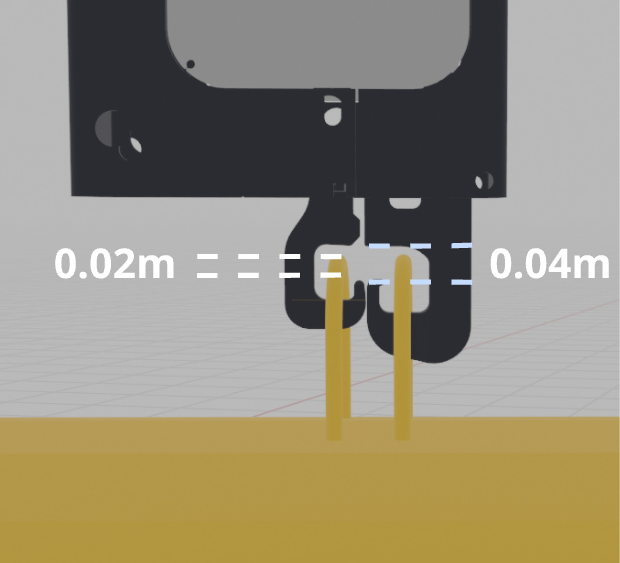}\\[6pt]
		\includegraphics[width=\linewidth]{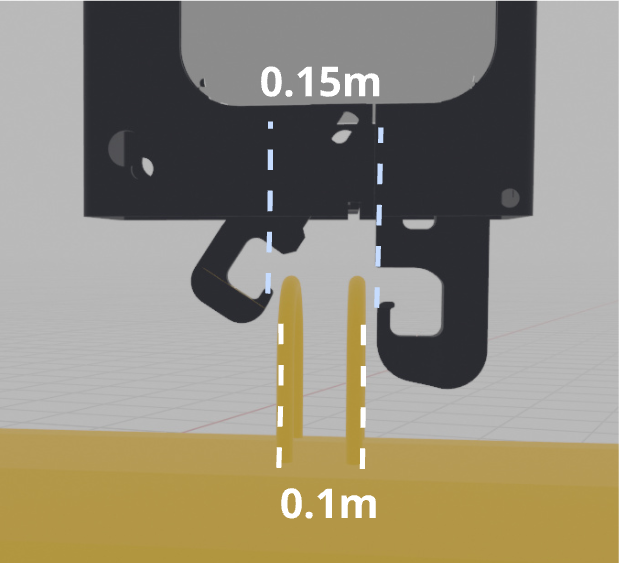}
	\end{minipage}
	\caption{System overview. \textbf{Left:} crane kinematic model and task setup. \textbf{Right:} close-ups illustrating tight hook--ring tolerances.}
	\label{fig:system_description}
\end{figure}

The task requires accurate TCP positioning and swing suppression due to tight geometric
tolerances between hooks and rings (Fig.~\ref{fig:system_description}). We assume an
external perception system provides accurate 3D poses; perception uncertainty is not
modeled.

\subsection{Residual Reinforcement Learning}
The task is simultaneous TCP trajectory tracking and swing suppression during hooking and
lifting. A trusted Cartesian TCP trajectory is generated from the initial TCP and container
poses and split into three segments (approach, horizontal alignment, lift), which introduced in Section~\ref{subsec:episode_initialization}. We adopt an RRL architecture (Fig.~\ref{fig:architecture}): a nominal controller provides stable baseline
tracking and anti-swing behavior, while a residual policy compensates unmodeled dynamics.

\subsubsection{Nominal Controller}
The nominal controller operates in Cartesian space at the TCP and consists of:
(i) an admittance controller for trajectory tracking,
(ii) a pendulum-aware anti-swing acceleration term,
and (iii) a damped least-squares inverse kinematics (IK) mapping to joint space.

\paragraph{\textbf{Admittance Control with Anti-Swing Compensation.}}
Let $x,v\in\mathbb{R}^3$ be TCP position/velocity and $x_{\mathrm{ref}},v_{\mathrm{ref}}$ the reference.
Define
\begin{equation}
	e_p=x_{\mathrm{ref}}-x,\qquad e_v=v_{\mathrm{ref}}-v, \qquad e_i(t) = \int_0^t e_p(\tau)\,\mathrm{d}\tau,
\end{equation}
We construct a \emph{virtual force} command
\begin{equation}
	F_{\mathrm{cmd}}=K_pe_p+K_ve_v+K_ie_i+M_d a_{xy},
\end{equation}
and integrate the admittance model
\begin{equation}
\begin{aligned}
	M_d \dot v_d + D_d v_d + K_d (x_d - x_{\mathrm{ref}}) = F_{\mathrm{cmd}},\\
	v_d^{k+1} = \mathrm{sat}_{v_\mathrm{max}}\!\left( v_d^k + \Delta t\, \dot{v}_d^k \right).
\end{aligned}
\end{equation}

\paragraph{\textbf{Pendulum-Aware Anti-Swing Acceleration.}}
For small swing angles, the discharge unit behaves approximately as
\begin{equation}
	L\ddot{\theta}+g\theta \approx -a_{\mathrm{TCP}},
\end{equation}
so horizontal TCP acceleration shapes sway~\cite{Sun2017OffshoreCrane,Qiang21}. We command
\begin{equation}
	a_{\mathrm{TCP}}=k_\theta\theta+k_\omega\dot\theta,
\end{equation}
yielding
\begin{equation}
	\ddot{\theta}+\frac{k_\omega}{L}\dot{\theta}+\frac{g+k_\theta}{L}\theta=0.
\end{equation}
Matching to the standard second-order form
\begin{equation}
	\ddot{\theta}+2\zeta\omega_n\dot{\theta}+\omega_n^2\theta=0,
\end{equation}
gives
\begin{equation}
	k_\theta(L)=L\omega_n^2-g,\qquad k_\omega(L)=2\zeta L\omega_n.
\end{equation}
We apply this in both horizontal directions and filter swing estimates:
\begin{equation}
	a_{xy}
	= w_s
	\begin{bmatrix}
		k_\theta \hat{\theta}_x + k_\omega \hat{\dot{\theta}}_x \\
		k_\theta \hat{\theta}_y + k_\omega \hat{\dot{\theta}}_y \\
		0
	\end{bmatrix},
	\quad
	\hat{\theta}^{k+1}_{xy} = (1-\alpha)\hat{\theta}^k_{xy} + \alpha\theta^k_{xy},
	\quad
	\hat{\dot{\theta}}^{k+1}_{xy} = (1-\alpha)\hat{\dot{\theta}}^k_{xy} + \alpha\dot{\theta}^k_{xy}.
\end{equation}

\paragraph{\textbf{Inverse Kinematics with Nullspace Optimization.}}
The desired TCP velocity $v_d$ is mapped to joint velocities using damped
least-squares inverse kinematics augmented with a nullspace posture term:
\begin{equation}
	{}^{\mathrm{nor}}u = J^+_\lambda v_d + (I - J^+_\lambda J)\, k_{\mathrm{ns}} (q_c - q),
	\label{eq:norminal_action}
\end{equation}

\subsubsection{Learning the Residual Action}
We follow the RRL formulation~\cite{johannink2018}. The residual policy outputs a correction
$a_r$ to improve precision under unmodeled dynamics. Since tight tolerances matter most during
horizontal alignment, we apply the residual only in segment~B (Fig.~\ref{fig:epis_Initialization}):
\[
u={}^{\mathrm{nor}}u  + {}^{\mathrm{res}}u \ \text{(segment B)}, \qquad u={}^{\mathrm{nor}}u \ \text{(segments A,C)}.
\]
We use a trajectory-tube metric~\cite{spinelli2025} to quantify adherence:
\begin{equation}
\delta_{\text{tube}}
=
\max\!\left(
-\frac{r}{2},\;
r - \frac{\|\left(\mathbf p_{\mathrm{tcp}} - \mathbf p_1\right)\times\left(\mathbf p_2 - \mathbf p_1\right)\|}{\|\mathbf p_2 - \mathbf p_1\|}
\right),
\label{eq:tube_delta}
\end{equation}
where $r$ is the tube radius and $\delta_{\text{tube}}>0$ indicates the TCP remains inside the tube.
The policy is trained with PPO using \texttt{skrl}~\cite{serrano2023skrl} in Isaac Lab; the MLP has hidden sizes $[128,64,32]$.

\subsubsection*{A. Observations and Actions}

The policy input consists of an 78-dimensional observation vector formed by concatenating task-relevant geometric information, and previous action, proprioceptive states history over the last three time steps (Table~\ref{tab:observation}). This history allows the policy to infer system dynamics and controller behavior. The tube distance $\delta_{\text{tube}}$ is included explicitly to inform the policy about trajectory adherence.

\renewcommand{\arraystretch}{1.3}
\begin{table}[h]
	\centering
	\begin{tabular}{lll}
		\hline
		\textbf{Observation} & \textbf{Notation} & \textbf{Dim.} \\
		\hline
		Crane joint positions & $q_{1:7}^{t:t-2}$ & 21 \\
		Crane joint velocities & $\dot q_{1:7}^{t:t-2}$ & 21 \\
		Discharge unit joint positions & $q_{10:11}^{t:t-2}$ & 6 \\
		Discharge unit joint velocities & $\dot q_{10:11}^{t:t-2}$ & 6 \\
		Reference TCP points & ${}^{\mathrm{rcs}}p^{t:t-2}$ & 9 \\
		Trajectory tube state & $\delta_{\text{tube}}$ & 1 \\
		Previous nominal action & ${}^{\mathrm{nor}}u_{1:7}^{t-1}$ & 7 \\
		Previous residual action & ${}^{\mathrm{res}}u_{1:7}^{t-1}$ & 7 \\
		\hline
	\end{tabular}
	\caption{Policy observations.}
	\label{tab:observation}
\end{table}

The residual action is a joint velocity correction for the crane:
\[
{}^{\mathrm{res}}u =
\left[
\hat{\dot q}_1,\ldots,\hat{\dot q}_7
\right].
\]

\subsubsection*{B. Reward and Termination}

The reward at time step $k$ is defined as a weighted sum of task-relevant components,
\begin{equation}
	\begin{aligned}
		R_k
		&= c_1 r^{\text{target\_coarse}}_k
		+ c_2 r^{\text{target\_fine}}_k
		+ c_3 r^{\text{tube}}_k
		+ c_4 r^{\text{progress}}_k \\
		&\quad
		+ c_7 r^{\text{oscillation}}_k
		+ c_8 r^{\text{lifting}}_k
		+ c_9 r^{\text{smooth}}_k ,
	\end{aligned}
\end{equation}
where $c_i$ are scalar weights.

The reward terms are computed using the TCP position $p^{\mathrm{tcp}}_k$, discharge unit position $p^d_k$, container position $p^c_k$, and the current reference control point $p^{\mathrm{ref}}_m$. 
The individual reward components are defined as
\begin{align*}
	r^{\text{target\_coarse}}_k
	&= -\frac{1}{\sigma}\max(0, d_{m,k}-\sigma),\\
	r^{\text{target\_fine}}_k
	&= 1-\tanh\!\left(\frac{d_{m,k}}{\sigma}\right),\\
	r^{\text{tube}}_k
	&=
	\mathbb{I}\!\Big[
	\delta_{\text{tube},k} \ge 0
	\wedge
	(p^{\mathrm{tcp}}_k-p^{\mathrm{ref}}_{m-1})^\top (p^{\mathrm{ref}}_m-p^{\mathrm{ref}}_{m-1}) \ge 0\\
	&\qquad\qquad\wedge
	(p^{\mathrm{tcp}}_k-p^{\mathrm{ref}}_m)^\top (p^{\mathrm{ref}}_m-p^{\mathrm{ref}}_{m-1}) \le 0
	\Big],\\
	r^{\text{progress}}_k
	&= m/M,\\
	r^{\text{oscillation}}_k
	&=
	1-\tanh\!\left(
	\dfrac{
		\arccos\!\left(\dfrac{\vec v_k^\top \mathbf g}{\|\vec v_k\|}\right)-\theta_{\max}
	}{\theta_{\max}}
	\right),\\
	r^{\text{lifting}}_k
	&= \mathbb{I}(z^c_k > z_{\min}),\\
	r^{\text{smooth}}_k
	&= -\sum_{i=1}^{N}\left(a^{\mathrm{res}}_{k,i}\right)^2 ,
\end{align*}
where
\[
d_{m,k}=\|p^{\mathrm{ref}}_m-p^{\mathrm{tcp}}_k\|_2,
\qquad
\vec v_k=p^d_k-p^{\mathrm{tcp}}_k,
\]

The reward terms have complementary roles. The tracking rewards
$r^{\text{target\_coarse}}_k$ and $r^{\text{target\_fine}}_k$ drive the TCP toward the
current control point, with coarse shaping far from the target and fine shaping near it.
$r^{\text{tube}}_k$ encourages the TCP to stay within a tubular neighborhood of the active
trajectory segment, while $r^{\text{progress}}_k$ promotes forward motion along the path.
$r^{\text{oscillation}}_k$ damps sway of the discharge unit relative to gravity, $r^{\text{lifting}}_k$ rewards successful lift, and $r^{\text{smooth}}_k$ penalizes large residual actions to encourage safe,
smooth control.

\begin{table}[h]
	\centering
	\renewcommand{\arraystretch}{1.8}
	\begin{tabular}{p{0.52\linewidth} p{0.42\linewidth}}
		\hline
		\textbf{Termination} & \textbf{Definition} \\
		\hline
		Distance out of range (current control point) &
		$\left\lVert \mathbf{p}^{\mathrm{ref}}_k - \mathbf{p}^{\mathrm{pb}}_k \right\rVert_2 > d_{\max}$ \\
		TCP outside trajectory tube for $n$ steps &
		$\sum_{j=1}^{k} \mathbb{I}\!\left(\delta^{\mathrm{tube}}_j < 0\right) \ge n$ \\
		Discharge unit gravity misalignment &
		$\arccos\!\left(\dfrac{(p^d_k - p^{\mathrm{tcp}}_k)^\top \mathbf{g}}{\lVert p^d_k - p^{\mathrm{tcp}}_k\rVert_2 \,\lVert \mathbf{g}\rVert_2}\right) > \theta_{\max}$ \\
		\hline
	\end{tabular}
	\vspace{0.1cm}
	\caption{Termination conditions.}
	\label{tab:termination}
\end{table}

To avoid unsafe or unrecoverable behaviors, episodes terminate early when
(i) the TCP deviates too far from the current reference point,
(ii) tube violations persist for $n$ times, or
(iii) the discharge unit tilt relative to gravity exceeds $\theta_{\max}$. The termination event are list in Table~\ref{tab:termination}.

\subsubsection*{C. Episode Initialization and Domain Randomization}
\label{subsec:episode_initialization}
\begin{figure}[t]
	\centering
	\includegraphics[width=0.8\textwidth]{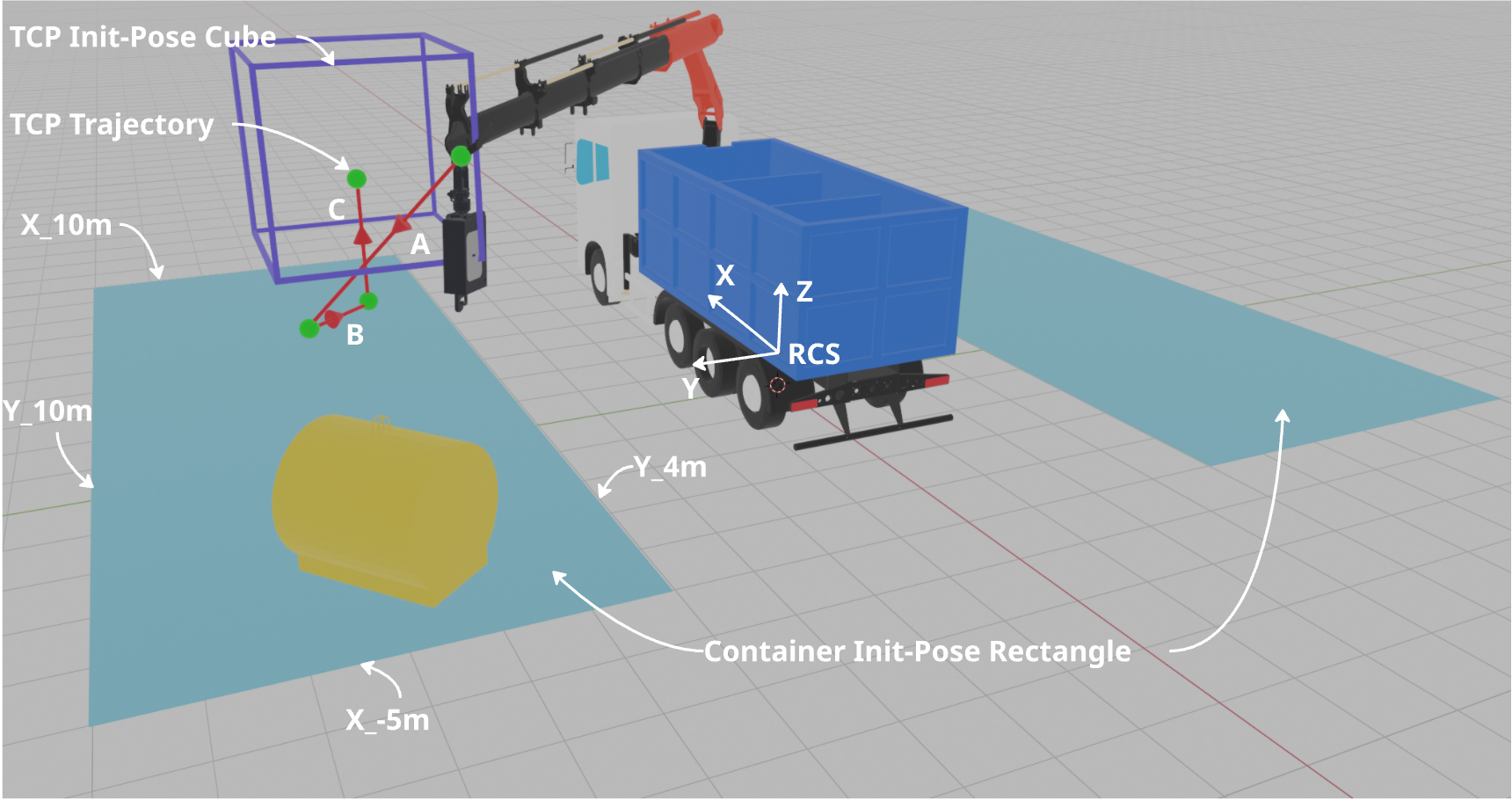}
	\caption{Randomized episode initialization: container spawned around the truck; TCP initialized in a cube above the container. TCP trajectory consists 3 segments (A, B, C) is generated according container and TCP position.}
	\label{fig:epis_Initialization}
\end{figure}

\paragraph{\textbf{Episode Initialization.}}
At reset, the container pose is uniformly sampled within a bounded horizontal workspace
around the truck (fixed height for ground contact) and its yaw is randomized within
preset limits. Conditioned on the sampled container pose, the crane TCP is initialized
by uniformly sampling a position in a cubic region above the container
(Fig.~\ref{fig:epis_Initialization}); the corresponding initial joint configuration is
computed via inverse kinematics. A reference TCP trajectory is then generated as a
spline between the initial and target poses and discretized into a fixed number of
control points that are provided sequentially as tracking targets.

\paragraph{\textbf{Domain Randomization.}}
To reduce sensitivity to modeling errors, we apply domain randomization at episode
reset. We perturb container mass and center of mass to vary payload and pendulum
dynamics, randomize actuator stiffness and damping to capture uncertainties in joint
behavior, and randomize friction parameters of the unactuated discharge-unit joints.
Finally, we scale the nominal admittance-controller gains to expose the residual policy
to a range of baseline controller behaviors.

\section{Experiments}

All experiments are conducted in simulation using Isaac Lab. The goal is to evaluate
the proposed residual reinforcement learning (RRL) framework on the container lifting
task with a large-scale loader crane and an underactuated discharge unit. Performance is
evaluated along four aspects:
(i) TCP trajectory tracking accuracy,
(ii) satisfaction of the trajectory tube constraint,
(iii) swing suppression of the discharge unit,
and (iv) robustness to parameter variations.
For unbiased evaluation, test episodes are evenly sampled from container initial
positions on the left and right sides of the truck workspace
(Fig.~\ref{fig:epis_Initialization}), and all metrics are aggregated over these
symmetric configurations.

\subsection{Trajectory Tracking and Tube Constraint}

\begin{figure}[t]
	\centering
	
	\begin{minipage}[t]{0.42\textwidth}
		\vspace{0pt}
		\centering
		\begin{subfigure}[t]{\linewidth}
			\centering
			\includegraphics[width=\linewidth]{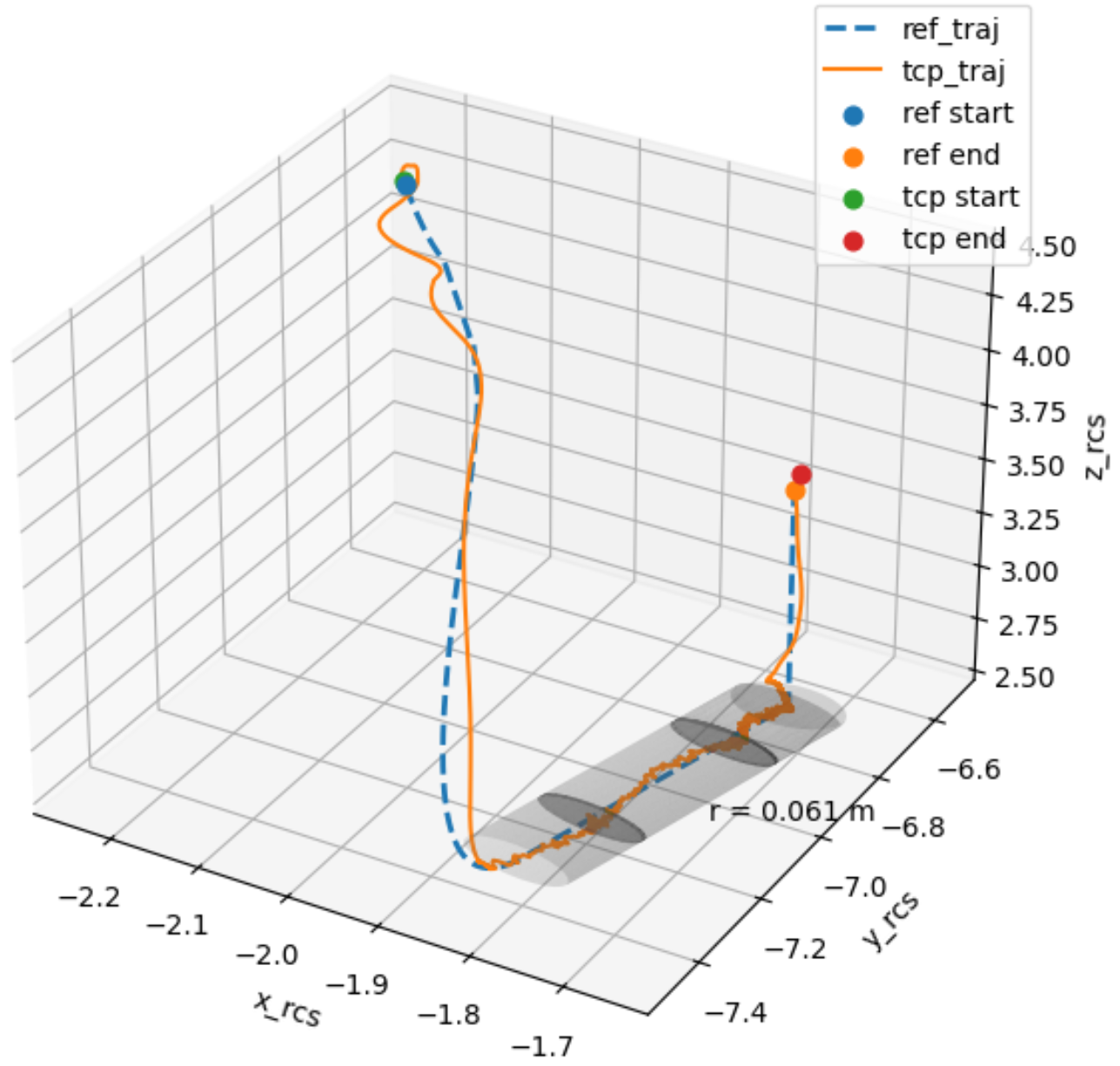}
			\caption{}
		\end{subfigure}
	\end{minipage}\hfill
	\begin{minipage}[t]{0.57\textwidth}
		\vspace{0pt}
		\centering
		
		\begin{subfigure}[t]{0.33\linewidth}
			\includegraphics[width=\linewidth]{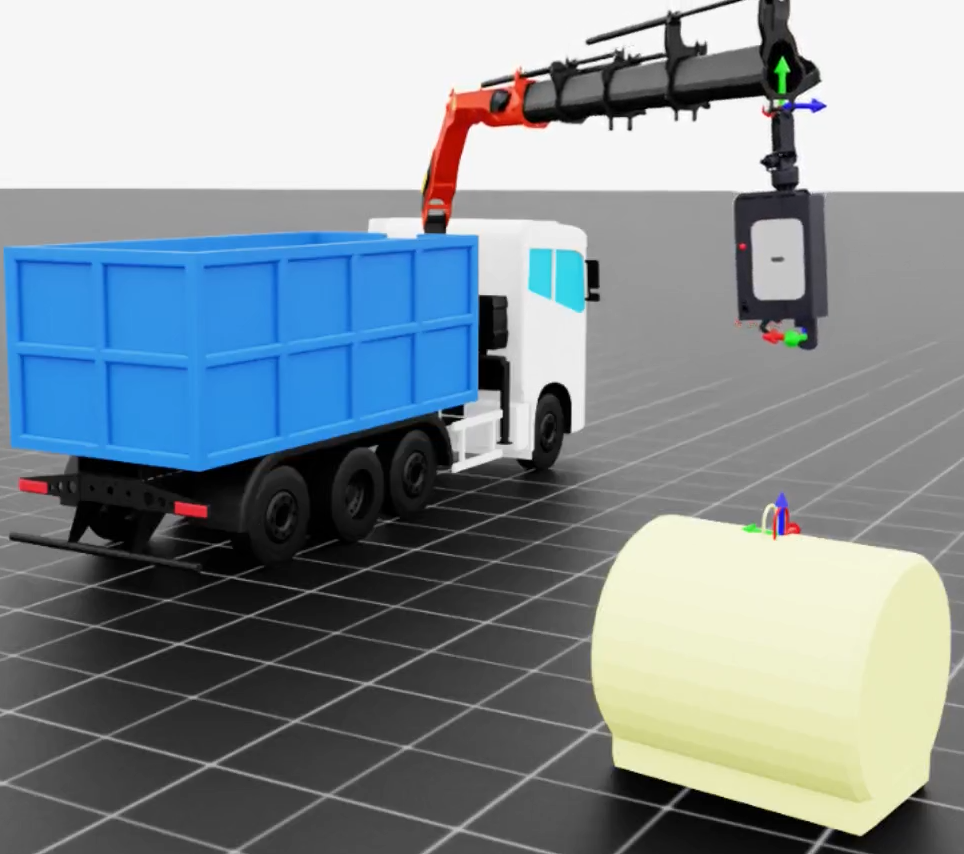}
			\caption{}
		\end{subfigure}\hfill
		\begin{subfigure}[t]{0.33\linewidth}
			\includegraphics[width=\linewidth]{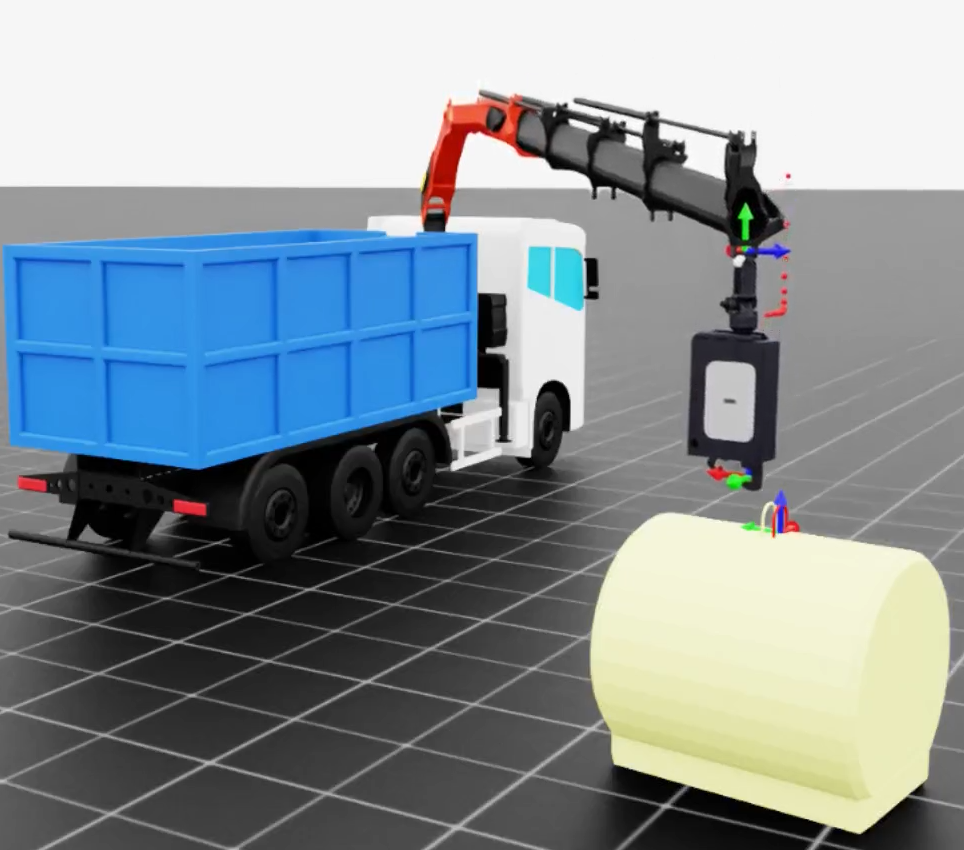}
			\caption{}
		\end{subfigure}\hfill
		\begin{subfigure}[t]{0.33\linewidth}
			\includegraphics[width=\linewidth]{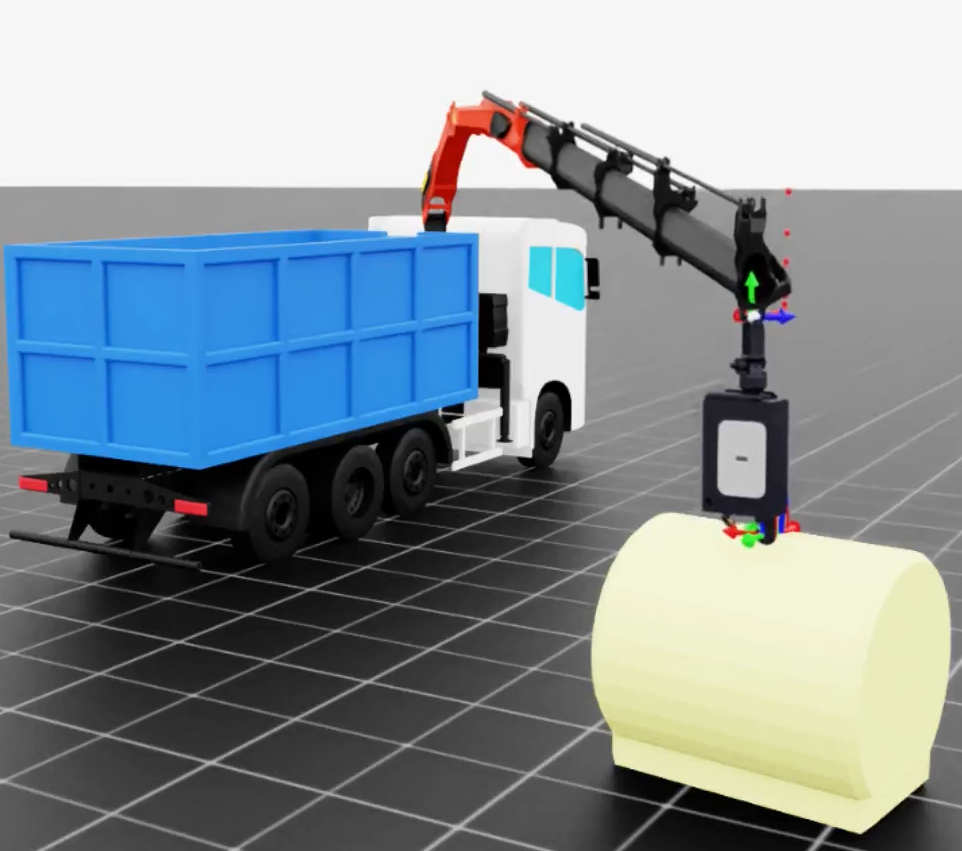}
			\caption{}
		\end{subfigure}
		
		\vspace{12pt}
		
		\begin{subfigure}[t]{0.33\linewidth}
			\includegraphics[width=\linewidth]{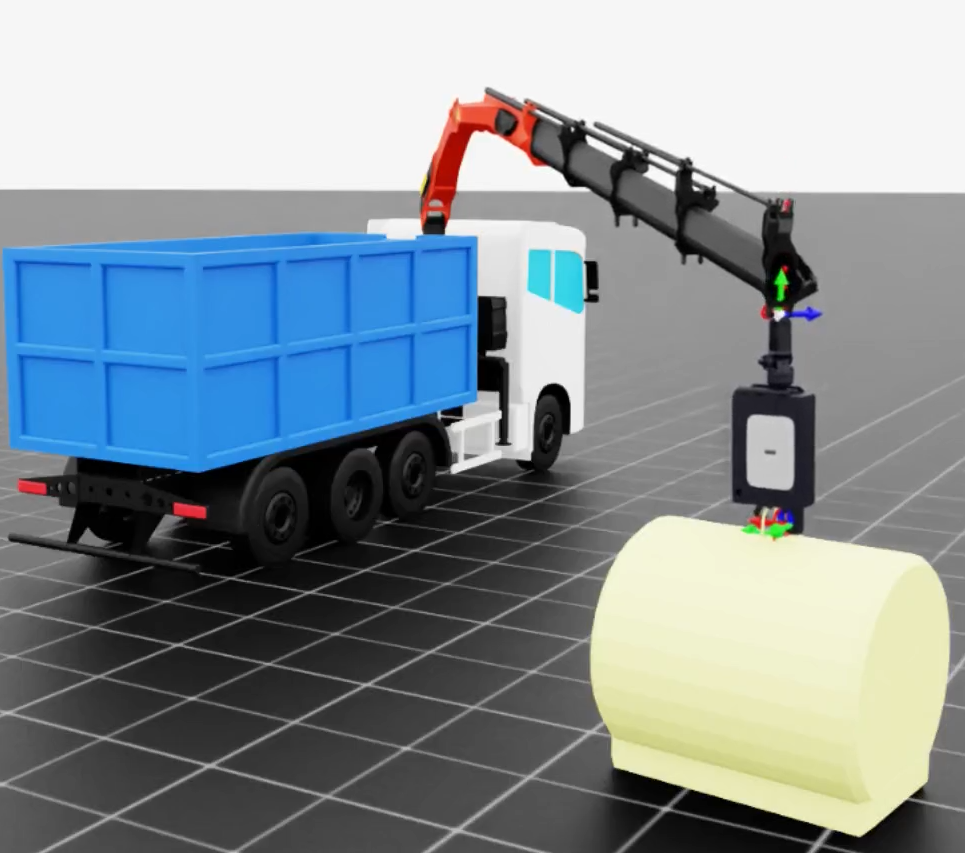}
			\caption{}
		\end{subfigure}\hfill
		\begin{subfigure}[t]{0.33\linewidth}
			\includegraphics[width=\linewidth]{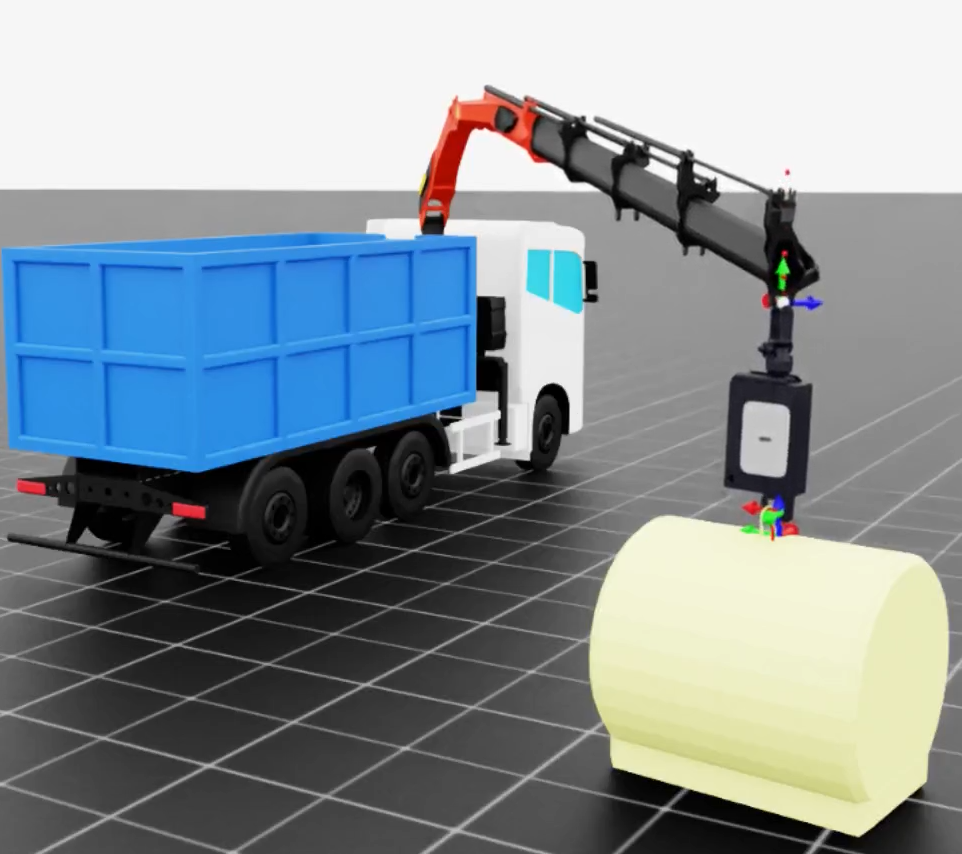}
			\caption{}
		\end{subfigure}\hfill
		\begin{subfigure}[t]{0.32\linewidth}
			\includegraphics[width=\linewidth]{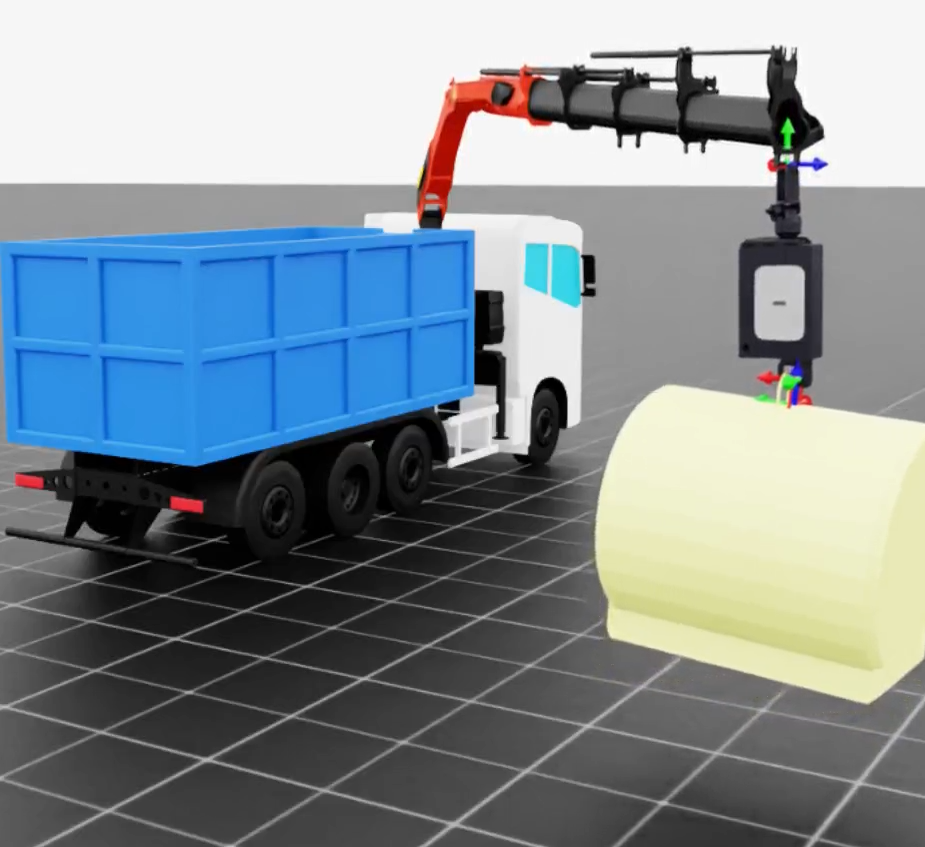}
			\caption{}
		\end{subfigure}
		
	\end{minipage}
	
	\caption{Representative lifting episode: (a) reference vs. executed TCP trajectory; (b--g) motion sequence.}
	\label{fig:traj}
\end{figure}

Figure~\ref{fig:traj}(a) illustrates a representative reference TCP trajectory and the
corresponding executed trajectory, while Fig.~\ref{fig:traj}(b--g) shows the crane motion
sequence. Tracking accuracy is quantified by the Euclidean distance between the executed
TCP and the reference trajectory.

\begin{figure}[t]
	\centering
	\includegraphics[width=1.05\textwidth]{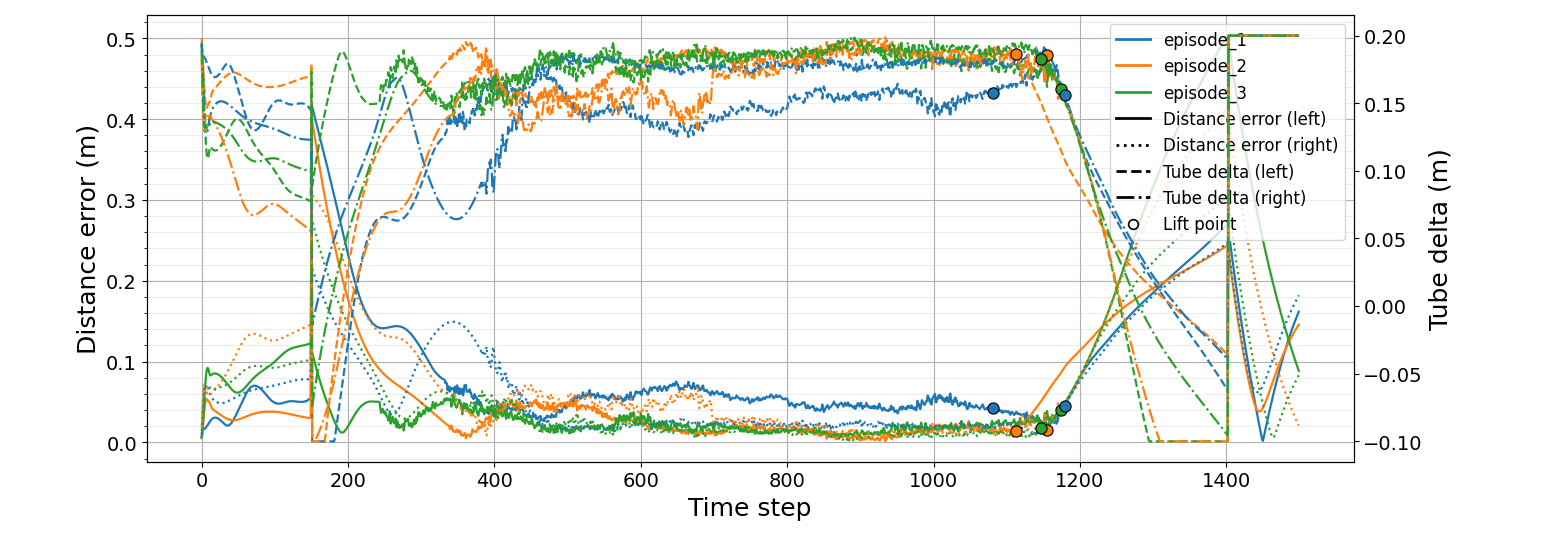}
	\caption{TCP tracking error and trajectory tube delta for representative episodes.}
	\label{fig:error_delta}
\end{figure}

To analyze accuracy relative to the trajectory tube constraint, we record both the TCP
tracking error and the tube delta (Eq.~\ref{eq:tube_delta}) over time. Figure~\ref{fig:error_delta} reports
results from three representative episodes on each side of the workspace. A larger tube
delta corresponds to the TCP remaining closer to the reference trajectory and is
consistently associated with lower tracking error. At the lifting instant, the TCP
tracking error remains below $0.04\,\mathrm{m}$, which is sufficient to satisfy the tight
geometric tolerances required for reliable container hooking.

\subsection{Swing Suppression}

\begin{figure}[t]
	\centering
	\includegraphics[width=1.0\textwidth]{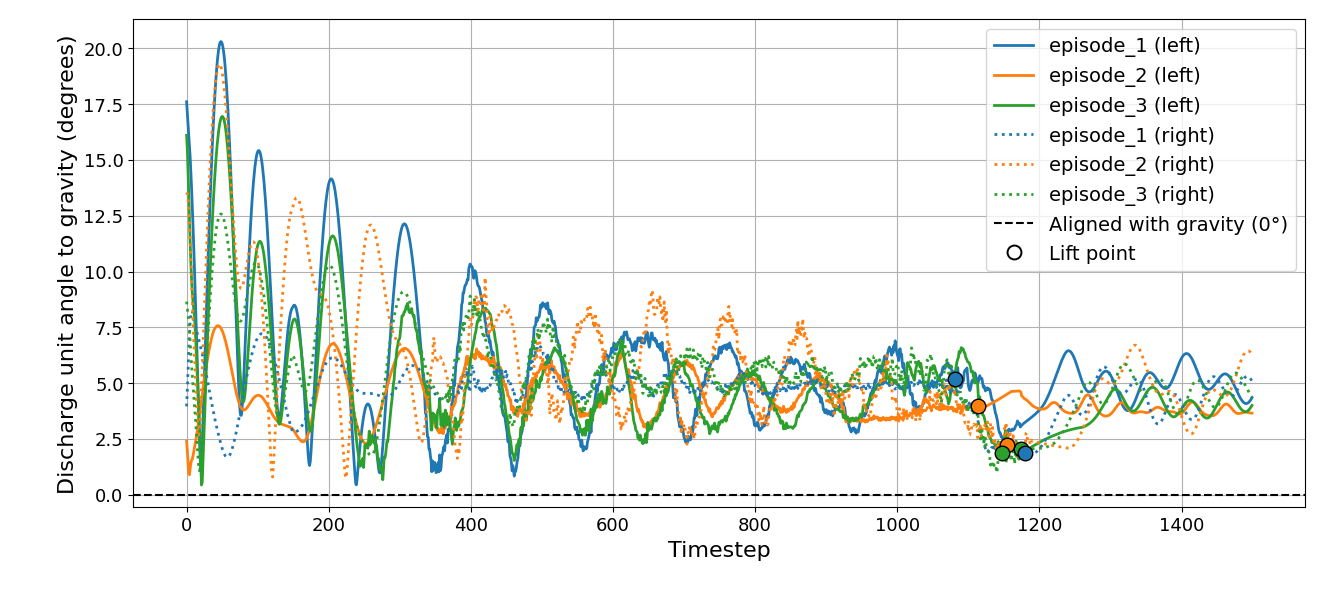}
	\caption{Swing angle of the discharge unit relative to gravity over time.}
	\label{fig:swing_angle}
\end{figure}

Figure~\ref{fig:swing_angle} shows the swing angle of the discharge unit relative to
gravity for the same evaluation episodes. Initial oscillations induced by crane motion
are effectively damped over time by the nominal controller and residual policy. At the
lifting point, four out of six episodes exhibit swing angles below $2.5^\circ$, while
all episodes remain within acceptable limits. Compared to peak initial swings of up to
$17.5^\circ$, this demonstrates substantial reduction in oscillation amplitude prior to
lifting.

\begin{table}[t]
	\centering
	\renewcommand{\arraystretch}{1.2}
	\setlength{\tabcolsep}{5pt}
	
	\begin{tabular}{lcccccccc}
		\toprule
		
		\multirow{3}{*}{\textbf{Randomization}}
		& \multicolumn{4}{c}{\textbf{Tracking Error(m)}}
		& \multicolumn{4}{c}{\textbf{Swing Angle(deg)}} \\
		\cmidrule(lr){2-5}\cmidrule(lr){6-9}
		& \multirow{2}{*}{\textbf{Mean}}
		& \multirow{2}{*}{\textbf{Std}}
		& \multirow{2}{*}{\textbf{\begin{tabular}[c]{@{}c@{}}Mean\\$_{[700,1200)}$\end{tabular}}}
		& \multirow{2}{*}{\textbf{\begin{tabular}[c]{@{}c@{}}Std\\$_{[700,1200)}$\end{tabular}}}
		& \multirow{2}{*}{\textbf{\begin{tabular}[c]{@{}c@{}}Mean\\@lift\end{tabular}}}
		& \multirow{2}{*}{\textbf{\begin{tabular}[c]{@{}c@{}}Std\\@lift\end{tabular}}}
		& \multirow{2}{*}{\textbf{\begin{tabular}[c]{@{}c@{}}Mean\\$_{[700,1200)}$\end{tabular}}}
		& \multirow{2}{*}{\textbf{\begin{tabular}[c]{@{}c@{}}Std\\$_{[700,1200)}$\end{tabular}}} \\
		& & & & & & & & \\
		\midrule
		scale\_0.1\_0.49  & 0.147 & 0.087 & 0.103 & 0.127 & 3.096 & 3.555 & 6.271 & 5.166 \\
		scale\_0.5\_1.5   & 0.073 & 0.019 & 0.026 & 0.015 & 2.046 & 1.316 & 4.016 & 1.378 \\
		scale\_1.51\_2.0  & 0.049 & 0.008 & 0.014 & 0.004 & 2.584 & 1.641 & 4.544 & 1.414 \\
		
		\addlinespace[6pt] 
		
		\midrule
		\multirow{3}{*}{\textbf{Randomization}}
		& \multicolumn{4}{c}{\textbf{Tube Delta(m)}}
		& \multicolumn{4}{c}{\textbf{Success Rate}} \\
		\cmidrule(lr){2-5}\cmidrule(lr){6-9}
		& \multirow{2}{*}{\textbf{Mean}}
		& \multirow{2}{*}{\textbf{Std}}
		& \multirow{2}{*}{\textbf{\begin{tabular}[c]{@{}c@{}}Mean\\$_{[700,1200)}$\end{tabular}}}
		& \multirow{2}{*}{\textbf{\begin{tabular}[c]{@{}c@{}}Std\\$_{[700,1200)}$\end{tabular}}}
		& \multicolumn{4}{c}{\multirow{2}{*}{\textbf{$z^c > 0.5\,\mathrm{m}$}}} \\
		& & & & & \multicolumn{4}{c}{} \\
		\midrule
		scale\_0.1\_0.49  & 0.089 & 0.050 & 0.114 & 0.077 & \multicolumn{4}{c}{47.3\%} \\
		scale\_0.5\_1.5   & 0.140 & 0.015 & 0.174 & 0.015 & \multicolumn{4}{c}{90.0\%} \\
		scale\_1.51\_2.0  & 0.158 & 0.008 & 0.187 & 0.004 & \multicolumn{4}{c}{92.3\%} \\
		\bottomrule
	\end{tabular}
	
	\caption{Performance metrics aggregated over three randomization groups.}
	\label{tab:robust_test}
\end{table}

\subsection{Robustness to Parameter Variations}

Robustness is evaluated under parameter settings that differ from those used during
training. During training, actuator gains, passive joint damping, and nominal controller
gains are randomized within scale of $[0.5,\,1.5]$. For evaluation, three disjoint ranges are
considered:
(i) scale\_0.1--0.49 (softer system),
(ii) scale\_0.5--1.5 (training range),
(iii) scale\_1.51--2.0 (stiffer system).
Each setting is evaluated over 300 episodes (150 left, 150 right). The results are
summarized in Table~\ref{tab:robust_test}.

The stiff configuration (scale\_1.51--2.0) achieves the best tracking accuracy and
highest success rate, as increased stiffness reduces crane bending and improves
trajectory following. However, swing angles are slightly larger due to altered crane
dynamics. The soft configuration (scale\_0.1--0.49) is the most challenging, resulting
in larger tracking errors and reduced tube adherence due to increased compliance.
Nevertheless, even in this case the method achieves a $47.3\%$ lifting success rate,
with mean swing angles at lift around $3^\circ$, which remains acceptable given the
length of the discharge unit.

Overall, these results demonstrate that the proposed RRL framework significantly improves
trajectory tracking, swing suppression, and robustness compared to the nominal controller
alone, particularly under challenging dynamic conditions.

\section{Ablation Studies}

To assess the contribution of each component in the proposed control architecture
(Fig.~\ref{fig:architecture}), we perform an ablation study on four controller variants:
(i) nominal controller with trajectory tracking only,
(ii) trajectory tracking with residual reinforcement learning (RRL),
(iii) trajectory tracking with anti-swing control,
and (iv) trajectory tracking with anti-swing control and RRL.
For configurations involving RRL, the residual policy is retrained accordingly.
Each variant is evaluated over 300 episodes (150 left, 150 right), and the results are
summarized in Table~\ref{tab:ablation_results}.

\paragraph{\textbf{Anti-Swing Control.}}
Without RRL, adding the anti-swing module improves the success rate from $57.3\%$ to
$71.3\%$, demonstrating its effectiveness in reducing pendulum oscillations and
improving task reliability.

\paragraph{\textbf{Residual Reinforcement Learning.}}
Introducing RRL significantly boosts performance in both cases.
Without anti-swing, RRL increases the success rate from $57.3\%$ to $88.3\%$,
showing its ability to compensate for unmodeled dynamics.
When anti-swing is present, RRL further improves the success rate from $71.3\%$ to
$91.7\%$, indicating complementary benefits.

\begin{table}[t]
	\centering
	\renewcommand{\arraystretch}{1.2}
	\setlength{\tabcolsep}{2pt}
	
	\begin{tabular}{ccccccccccc}
		\toprule
		
		\multicolumn{3}{c}{\textbf{Models}} & \multicolumn{4}{c}{\textbf{Tracking Error(m)}} & \multicolumn{4}{c}{\textbf{Swing Angle(deg)}} \\
		\cmidrule(lr){1-3}\cmidrule(lr){4-7}\cmidrule(lr){8-11}
		\multicolumn{2}{c}{\textbf{Nominal Controller}} & \multirow{2}{*}{\textbf{RRL}} & \multirow{2}{*}{\textbf{Mean}} & \multirow{2}{*}{\textbf{Std}} & \multirow{2}{*}{\textbf{\begin{tabular}[c]{@{}c@{}}Mean\\$_{[700,1200)}$\end{tabular}}} & \multirow{2}{*}{\textbf{\begin{tabular}[c]{@{}c@{}}Std\\$_{[700,1200)}$\end{tabular}}} & \multirow{2}{*}{\textbf{\begin{tabular}[c]{@{}c@{}}Mean\\@lift\end{tabular}}} & \multirow{2}{*}{\textbf{\begin{tabular}[c]{@{}c@{}}Std\\@lift\end{tabular}}} & \multirow{2}{*}{\textbf{\begin{tabular}[c]{@{}c@{}}Mean\\$_{[700,1200)}$\end{tabular}}} & \multirow{2}{*}{\textbf{\begin{tabular}[c]{@{}c@{}}Std\\$_{[700,1200)}$\end{tabular}}} \\
		\cmidrule(lr){1-2}
		\textbf{Traj. Tracking} & \textbf{Anti-Swing} &  & & & & & & & & \\
		\midrule
		$\bullet$ & $-$ & $-$ & 0.100 & 0.025 & 0.089 & 0.030 & 3.318 & 3.100 & 7.792 & 2.819 \\
		$\bullet$ & $-$ & $\bullet$ & 0.071 & 0.018 & 0.021 & 0.014 & 2.361 & 1.520 & 4.355 & 1.533 \\
		$\bullet$ & $\bullet$ & $-$ & 0.105 & 0.031 & 0.098 & 0.039 & 2.719 & 2.353 & 6.545 & 2.880 \\
		$\bullet$ & $\bullet$ & $\bullet$ & 0.073 & 0.021 & 0.027 & 0.014 & 2.090 & 1.440 & 4.033 & 1.303 \\
		
		\addlinespace[6pt]
		\midrule
		\multicolumn{3}{c}{\textbf{Models}} & \multicolumn{4}{c}{\textbf{Tube Delta(m)}} & \multicolumn{4}{c}{\textbf{Success Rate}} \\
		\cmidrule(lr){1-3}\cmidrule(lr){4-7}\cmidrule(lr){8-11}
		\multicolumn{2}{c}{\textbf{Nominal Controller}} &  \multirow{2}{*}{\textbf{RRL}} & \multirow{2}{*}{\textbf{Mean}} & \multirow{2}{*}{\textbf{Std}} & \multirow{2}{*}{\textbf{\begin{tabular}[c]{@{}c@{}}Mean\\$_{[700,1200)}$\end{tabular}}} & \multirow{2}{*}{\textbf{\begin{tabular}[c]{@{}c@{}}Std\\$_{[700,1200)}$\end{tabular}}} & \multicolumn{4}{c}{\multirow{2}{*}{\textbf{$z^c > 0.5\,\mathrm{m}$}}}\\
		\cmidrule(lr){1-2}
		\textbf{Traj. Tracking} & \textbf{Anti-Swing} &  & & & & & & & & \\
		\midrule
		$\bullet$ & $-$ & $-$ & 0.110 & 0.023 & 0.111 & 0.030 & \multicolumn{4}{c}{57.3\%} \\
		$\bullet$ & $-$ & $\bullet$ & 0.140 & 0.014 & 0.179 & 0.014 & \multicolumn{4}{c}{88.3\%} \\
		$\bullet$ & $\bullet$ & $-$ & 0.105 & 0.027 & 0.103 & 0.039 & \multicolumn{4}{c}{71.3\%} \\
		$\bullet$ & $\bullet$ & $\bullet$ & 0.139 & 0.016 & 0.173 & 0.014 & \multicolumn{4}{c}{91.7\%} \\
		\bottomrule
	\end{tabular}
	
	\caption{Performance metrics aggregated over 4 different models.}
	\label{tab:ablation_results}
\end{table}

\paragraph{\textbf{Tracking and Swing Performance.}}
TCP tracking error and tube delta vary only slightly across configurations, confirming
that trajectory tracking is mainly handled by the nominal controller, with RRL providing
a modest refinement.
In contrast, swing angle metrics show substantial differences: anti-swing control
significantly reduces oscillations, and RRL further lowers swing amplitudes, achieving
the best overall performance when both are combined.

In summary, the ablation study confirms that the anti-swing module provides an effective
model-based foundation, while the residual reinforcement learning policy substantially
enhances robustness and success by compensating for unmodeled effects.

\section{Conclusion and Future Work}
This work presented a residual reinforcement learning approach for accurate container
lifting with a large-scale hydraulic loader crane and an underactuated discharge unit.
By combining a nominal admittance controller with pendulum-aware anti-swing control and
a learned residual policy, the proposed method achieves improved tracking accuracy,
effective swing suppression, and robust performance under parameter variations in
simulation.

Several directions are planned for future research. First, we will extend the framework
to explicitly account for hydraulic dynamics, enabling adaptation of the residual policy
to real actuator behavior. Second, more realistic simulation of hydraulic systems will
be incorporated during training to reduce the sim-to-real gap. Finally, we plan to
validate the proposed method on a real loader crane platform and investigate additional
domain adaptation techniques to further improve transfer from simulation to real-world
deployment.

\section*{Acknowledgements}
This work was carried out within a publicly funded project of the Commercial Vehicle
Cluster (CVC), Rheinland-Pfalz, Germany, in collaboration with Palfinger AG. Palfinger AG
provided the CAD model of the loader crane and the discharge unit used in this research
\printbibliography[heading=subbibintoc]

@inproceedings{Spinelli_2024,
	title={Reinforcement Learning Control for Autonomous Hydraulic Material Handling Machines with Underactuated Tools},
	url={http://dx.doi.org/10.1109/IROS58592.2024.10802199},
	DOI={10.1109/iros58592.2024.10802199},
	booktitle={2024 IEEE/RSJ International Conference on Intelligent Robots and Systems (IROS)},
	publisher={IEEE},
	author={Spinelli, Filippo A. and Egli, Pascal and Nubert, Julian and Nan, Fang and Bleumer, Thilo and Goegler, Patrick and Brockes, Stephan and Hofmann, Ferdinand and Hutter, Marco},
	year={2024},
	month=oct, pages={12694--12701} }

@misc{spinelli2025,
	title={Large Scale Robotic Material Handling: Learning, Planning, and Control}, 
	author={Filippo A. Spinelli and Yifan Zhai and Fang Nan and Pascal Egli and Julian Nubert and Thilo Bleumer and Lukas Miller and Ferdinand Hofmann and Marco Hutter},
	year={2025},
	eprint={2508.09003},
	archivePrefix={arXiv},
	primaryClass={cs.RO},
	url={https://arxiv.org/abs/2508.09003}
}

@ARTICLE{Quang02,
	author={Quang Ha and Santos, M. and Quang Nguyen and Rye, D. and Durrant-Whyte, H.},
	journal={IEEE Robotics \& Automation Magazine}, 
	title={Robotic excavation in construction automation}, 
	year={2002},
	volume={9},
	number={1},
	pages={20--28},
	doi={10.1109/100.993151}}

@article{CHANG2002119,
	title = {A straight-line motion tracking control of hydraulic excavator system},
	journal = {Mechatronics},
	volume = {12},
	number = {1},
	pages = {119--138},
	year = {2002},
	issn = {0957-4158},
	doi = {https://doi.org/10.1016/S0957-4158(01)00014-9},
	url = {https://www.sciencedirect.com/science/article/pii/S0957415801000149},
	author = {Pyung Hun Chang and Soo-Jin Lee},
}

@ARTICLE{Mattila17,
	author={Mattila, Jouni and Koivumäki, Janne and Caldwell, Darwin G. and Semini, Claudio},
	journal={IEEE/ASME Transactions on Mechatronics}, 
	title={A Survey on Control of Hydraulic Robotic Manipulators With Projection to Future Trends}, 
	year={2017},
	volume={22},
	number={2},
	pages={669--680},
	doi={10.1109/TMECH.2017.2668604}}

@article{Jud_2021,
	title={HEAP - The autonomous walking excavator},
	volume={129},
	ISSN={0926-5805},
	url={http://dx.doi.org/10.1016/j.autcon.2021.103783},
	DOI={10.1016/j.autcon.2021.103783},
	journal={Automation in Construction},
	publisher={Elsevier BV},
	author={Jud, Dominic and Kerscher, Simon and Wermelinger, Martin and Jelavic, Edo and Egli, Pascal and Leemann, Philipp and Hottiger, Gabriel and Hutter, Marco},
	year={2021},
	month=sep, pages={103783} }

@ARTICLE{Egli24,
	author={Egli, Pascal and Terenzi, Lorenzo and Hutter, Marco},
	journal={IEEE Transactions on Field Robotics}, 
	title={Reinforcement Learning-Based Bucket Filling for Autonomous Excavation}, 
	year={2024},
	volume={1},
	number={},
	pages={170--191},
	doi={10.1109/TFR.2024.3432508}}

@ARTICLE{Egli22,
	author={Egli, Pascal and Gaschen, Dominique and Kerscher, Simon and Jud, Dominic and Hutter, Marco},
	journal={IEEE Robotics and Automation Letters}, 
	title={Soil-Adaptive Excavation Using Reinforcement Learning}, 
	year={2022},
	volume={7},
	number={4},
	pages={9778--9785},
	doi={10.1109/LRA.2022.3189834}}

@misc{zhai2025,
	title={ExT: Towards Scalable Autonomous Excavation via Large-Scale Multi-Task Pretraining and Fine-Tuning}, 
	author={Yifan Zhai and Lorenzo Terenzi and Patrick Frey and Diego Garcia Soto and Pascal Egli and Marco Hutter},
	year={2025},
	eprint={2509.14992},
	archivePrefix={arXiv},
	primaryClass={cs.RO},
	url={https://arxiv.org/abs/2509.14992}, 
}

@misc{gruetter2025,
	title={Towards Learning Boulder Excavation with Hydraulic Excavators}, 
	author={Jonas Gruetter and Lorenzo Terenzi and Pascal Egli and Marco Hutter},
	year={2025},
	eprint={2509.17683},
	archivePrefix={arXiv},
	primaryClass={cs.RO},
	url={https://arxiv.org/abs/2509.17683}, 
}

@INPROCEEDINGS{Egli20,
	author={Egli, Pascal and Hutter, Marco},
	booktitle={2020 IEEE/RSJ International Conference on Intelligent Robots and Systems (IROS)}, 
	title={Towards RL-Based Hydraulic Excavator Automation}, 
	year={2020},
	volume={},
	number={},
	pages={2692--2697},
	doi={10.1109/IROS45743.2020.9341598}}

@ARTICLE{Wu24,
	author={Wu, Qingxiang and Sun, Ning and Yang, Tong and Fang, Yongchun},
	journal={IEEE Transactions on Industrial Electronics}, 
	title={Deep Reinforcement Learning-Based Control for Asynchronous Motor-Actuated Triple Pendulum Crane Systems With Distributed Mass Payloads}, 
	year={2024},
	volume={71},
	number={2},
	pages={1853--1862},
	doi={10.1109/TIE.2023.3262891}}

@misc{werner2024,
	title={Dynamic Throwing with Robotic Material Handling Machines}, 
	author={Lennart Werner and Fang Nan and Pol Eyschen and Filippo A. Spinelli and Hongyi Yang and Marco Hutter},
	year={2024},
	eprint={2405.19001},
	archivePrefix={arXiv},
	primaryClass={cs.RO},
	url={https://arxiv.org/abs/2405.19001}, 
}

@misc{andersson2021,
	title={Reinforcement Learning Control of a Forestry Crane Manipulator}, 
	author={Jennifer Andersson and Kenneth Bodin and Daniel Lindmark and Martin Servin and Erik Wallin},
	year={2021},
	eprint={2103.02315},
	archivePrefix={arXiv},
	primaryClass={cs.RO},
	url={https://arxiv.org/abs/2103.02315}, 
}

@article{Kulkarni22,
	author = {Kulkarni, Padmaja and Kober, Jens and Babuška, Robert and Della Santina, Cosimo},
	title = {Learning Assembly Tasks in a Few Minutes by Combining Impedance Control and Residual Recurrent Reinforcement Learning},
	journal = {Advanced Intelligent Systems},
	volume = {4},
	number = {1},
	pages = {2100095},
	doi = {https://doi.org/10.1002/aisy.202100095},
	url = {https://advanced.onlinelibrary.wiley.com/doi/abs/10.1002/aisy.202100095},
	eprint = {https://advanced.onlinelibrary.wiley.com/doi/pdf/10.1002/aisy.202100095},
	year = {2022}
}

@misc{alakuijala2021,
	title={Residual Reinforcement Learning from Demonstrations}, 
	author={Minttu Alakuijala and Gabriel Dulac-Arnold and Julien Mairal and Jean Ponce and Cordelia Schmid},
	year={2021},
	eprint={2106.08050},
	archivePrefix={arXiv},
	primaryClass={cs.LG},
	url={https://arxiv.org/abs/2106.08050}, 
}

@misc{ankile2025,
	title={Residual Off-Policy RL for Finetuning Behavior Cloning Policies}, 
	author={Lars Ankile and Zhenyu Jiang and Rocky Duan and Guanya Shi and Pieter Abbeel and Anusha Nagabandi},
	year={2025},
	eprint={2509.19301},
	archivePrefix={arXiv},
	primaryClass={cs.RO},
	url={https://arxiv.org/abs/2509.19301}, 
}

@misc{wallin2024,
	title={Multi-log grasping using reinforcement learning and virtual visual servoing}, 
	author={Erik Wallin and Viktor Wiberg and Martin Servin},
	year={2024},
	eprint={2309.02997},
	archivePrefix={arXiv},
	primaryClass={cs.RO},
	url={https://arxiv.org/abs/2309.02997}, 
}

@misc{nvidia2025,
	title={Isaac Lab: A GPU-Accelerated Simulation Framework for Multi-Modal Robot Learning}, 
	author={NVIDIA and Mayank Mittal and Pascal Roth and James Tigue and Antoine Richard and Octi Zhang and Peter Du and Antonio Serrano-Muñoz and Xinjie Yao and René Zurbrügg and Nikita Rudin and Lukasz Wawrzyniak and Milad Rakhsha and Alain Denzler and Eric Heiden and Ales Borovicka and Ossama Ahmed and Iretiayo Akinola and Abrar Anwar and Mark T. Carlson and Ji Yuan Feng and Animesh Garg and Renato Gasoto and Lionel Gulich and Yijie Guo and M. Gussert and Alex Hansen and Mihir Kulkarni and Chenran Li and Wei Liu and Viktor Makoviychuk and Grzegorz Malczyk and Hammad Mazhar and Masoud Moghani and Adithyavairavan Murali and Michael Noseworthy and Alexander Poddubny and Nathan Ratliff and Welf Rehberg and Clemens Schwarke and Ritvik Singh and James Latham Smith and Bingjie Tang and Ruchik Thaker and Matthew Trepte and Karl Van Wyk and Fangzhou Yu and Alex Millane and Vikram Ramasamy and Remo Steiner and Sangeeta Subramanian and Clemens Volk and CY Chen and Neel Jawale and Ashwin Varghese Kuruttukulam and Michael A. Lin and Ajay Mandlekar and Karsten Patzwaldt and John Welsh and Huihua Zhao and Fatima Anes and Jean-Francois Lafleche and Nicolas Moënne-Loccoz and Soowan Park and Rob Stepinski and Dirk Van Gelder and Chris Amevor and Jan Carius and Jumyung Chang and Anka He Chen and Pablo de Heras Ciechomski and Gilles Daviet and Mohammad Mohajerani and Julia von Muralt and Viktor Reutskyy and Michael Sauter and Simon Schirm and Eric L. Shi and Pierre Terdiman and Kenny Vilella and Tobias Widmer and Gordon Yeoman and Tiffany Chen and Sergey Grizan and Cathy Li and Lotus Li and Connor Smith and Rafael Wiltz and Kostas Alexis and Yan Chang and David Chu and Linxi "Jim" Fan and Farbod Farshidian and Ankur Handa and Spencer Huang and Marco Hutter and Yashraj Narang and Soha Pouya and Shiwei Sheng and Yuke Zhu and Miles Macklin and Adam Moravanszky and Philipp Reist and Yunrong Guo and David Hoeller and Gavriel State},
	year={2025},
	eprint={2511.04831},
	archivePrefix={arXiv},
	primaryClass={cs.RO},
	url={https://arxiv.org/abs/2511.04831}, 
}

@article{Sun2017OffshoreCrane,
	author  = {Sun, You-Gang and Qiang, Hai-Yan and Xu, Junqi and Dong, Da-Shan},
	title   = {The Nonlinear Dynamics and Anti-Sway Tracking Control for Offshore Container Crane on a Mobile Harbor},
	journal = {Journal of Marine Science and Technology},
	volume  = {25},
	number  = {6},
	year    = {2017},
	pages   = {Article 5},
	doi     = {10.6119/JMST-017-1226-05},
	url     = {https://jmstt.ntou.edu.tw/journal/vol25/iss6/5}
}

@ARTICLE{Qiang21,
	AUTHOR={Qiang, Hai-yan  and Sun, You-gang  and Lyu, Jin-chao  and Dong, Da-shan },
	TITLE={Anti-Sway and Positioning Adaptive Control of a Double-Pendulum Effect Crane System With Neural Network Compensation},
	JOURNAL={Frontiers in Robotics and AI},
	VOLUME={Volume 8 - 2021},
	YEAR={2021},
	URL={https://www.frontiersin.org/journals/robotics-and-ai/articles/10.3389/frobt.2021.639734},
	DOI={10.3389/frobt.2021.639734},
	ISSN={2296-9144}
	}

@article{Ryan23,
	author  = {Johns, Ryan Luke and Wermelinger, Martin and Mascaro, Ruben and Jud, Dominic and Hurkxkens, Ilmar and Vasey, Lauren and Chli, Margarita and Gramazio, Fabio and Kohler, Matthias and Hutter, Marco},
	title   = {A framework for robotic excavation and dry stone construction using on-site materials},
	journal = {Science Robotics},
	volume  = {8},
	number  = {84},
	pages   = {eabp9758},
	year    = {2023},
	doi     = {10.1126/scirobotics.abp9758},
	url     = {https://www.science.org/doi/10.1126/scirobotics.abp9758}
}

@article{han2023systematic,
	title   = {A systematic trajectory tracking framework for robot manipulators: An observer-based nonsmooth control approach},
	author  = {Han, Linyan and Mao, Jianliang and Zhang, Chuanlin and Kay, Robert W. and Richardson, Robert C. and Zhou, Chengxu},
	journal = {IEEE Transactions on Industrial Electronics},
	volume  = {71},
	number  = {9},
	pages   = {11104--11114},
	year    = {2023}
}

@misc{johannink2018,
	title        = {Residual Reinforcement Learning for Robot Control},
	author       = {Johannink, Tobias and Bahl, Shikhar and Nair, Ashvin and Luo, Jianlan
	and Kumar, Avinash and Loskyll, Matthias and Aparicio Ojea, Juan
	and Solowjow, Eugen and Levine, Sergey},
	year         = {2018},
	eprint       = {1812.03201},
	eprinttype   = {arxiv},
	eprintclass  = {cs.RO},
	url          = {https://arxiv.org/abs/1812.03201}
}

@article{serrano2023skrl,
	author  = {Antonio Serrano-Muñoz and Dimitrios Chrysostomou and Simon Bøgh and Nestor Arana-Arexolaleiba},
	title   = {skrl: Modular and Flexible Library for Reinforcement Learning},
	journal = {Journal of Machine Learning Research},
	year    = {2023},
	volume  = {24},
	number  = {254},
	pages   = {1--9},
	url     = {http://jmlr.org/papers/v24/23-0112.html}
}

\end{document}